\documentclass[10pt,conference]{IEEEtran}
\IEEEoverridecommandlockouts
\usepackage{cite}
\usepackage{amsmath,amssymb,amsfonts}
\usepackage{algorithmic}
\usepackage{graphicx}
\usepackage{textcomp}
\usepackage{xcolor}
\usepackage{xspace}
\usepackage{balance}
\usepackage{multicol}
\usepackage{hyperref}
\def\BibTeX{{\rm B\kern-.05em{\sc i\kern-.025em b}\kern-.08em
    T\kern-.1667em\lower.7ex\hbox{E}\kern-.125emX}}

\newcommand{\fontys}{Fontys University of Applied Sciences\xspace}
\newcommand{\flowerpower}{FlowerPower\xspace}
\newcommand{\flowerpowersystem}{FlowerPower  platform\xspace}

\begin{document}

\title{Defining Quality Requirements for a Trustworthy AI Wildflower Monitoring Platform
}

\author{\IEEEauthorblockN{Petra Heck}
\IEEEauthorblockA{
\textit{Fontys University of Applied Science}\\
Eindhoven, The Netherlands \\
p.heck@fontys.nl}
\and
\IEEEauthorblockN{Gerard Schouten}
\IEEEauthorblockA{
\textit{Fontys University of Applied Science}\\
Eindhoven, The Netherlands \\
g.schouten@fontys.nl}
\IEEEauthorblockA{
\textit{Naturalis Biodiversity Center}\\
Leiden, The Netherlands \\
gerard.schouten@naturalis.nl}
}

\maketitle

\begin{abstract}
For an AI solution to evolve from a trained machine learning model into a production-ready AI system, many more things need to be considered than just the performance of the machine learning model. A production-ready AI system needs to be trustworthy, i.e. of high quality. But how to determine this in practice? For traditional software, ISO25000 and its predecessors have since long time been used to define and measure quality characteristics. Recently, quality models for AI systems, based on ISO25000, have been introduced. This paper applies one such quality model to a real-life case study: a deep learning platform for monitoring wildflowers. The paper presents three realistic scenarios sketching what it means to respectively use, extend and incrementally improve the deep learning platform for wildflower identification and counting. Next, it is shown how the quality model can be used as a structured dictionary to define quality requirements for data, model and software. Future work remains to extend the quality model with metrics, tools and best practices to aid AI engineering practitioners in implementing trustworthy AI systems.  
\end{abstract}

\begin{IEEEkeywords}
software product quality, trustworthy AI, quality requirements, biodiversity monitoring, wildflowers
\end{IEEEkeywords}

\section{Introduction}
Biodiversity decline has life-impacting consequences \cite{ipbes}. Planet health is at stake, a major concern for both researchers and policy makers. They are in need of reliable tools and standards to monitor biodiversity. With recent advances in deep learning technology, it is possible to create scalable AI-enabled solutions for biodiversity monitoring \cite{klein}.

This paper describes a case study on the expansion of a prototype AI model for identifying and counting wildflowers to a full-fledged AI-enabled wildflower monitoring platform (called ``\href{https://fontys.nl/Onderzoek/AI-en-big-data/Flower-Power-AI-als-middel-om-biodiversiteit-te-meten.htm}{\flowerpower}'').

Recent advances in AI engineering \cite{Bosch2021} show that for an AI model to be used in practice, software engineering and data engineering is needed to embed the model into an AI-enabled software system. When deploying an AI system, one should also consider the effects this might have on the humans interacting with the system. People should be able to trust the AI systems they are working with \cite{EU}. 

Building trustworthy AI systems is not just about incorporating an AI model with high performance. In essence an AI system is a software product and thus its trustworthiness should also be considered from a software quality perspective. At the same time AI models are trained with (labeled) datasets. Thus the quality of the training data also has impact on the trustworthiness of the AI system.

This paper demonstrates for the concrete example of \flowerpower how to define trustworthiness for an AI system. The focus thereby is on the AI-specific aspects of the wildflower monitoring platform. The approach in this paper is also applicable to other AI-enabled software systems. 

Section \ref{sec:back} introduces background and related work on software quality, trustworthy AI and biodiversity monitoring. Section \ref{sec:case} describes the \flowerpowersystem in more detail. Section \ref{sec:req} analyzes the AI-related quality requirements that need to be fulfilled to make the \flowerpowersystem a trustworthy AI system. Section \ref{sec:discuss} discusses the future of \flowerpower and implications for other AI-enabled systems. The paper concludes with contributions and future work. 

\section{Background and Related Work}\label{sec:back}
This section summarizes the background and related work on software product quality, requirements and their application to AI systems. This section also gives an overview of how AI has been applied in the area of biodiversity monitoring until now. 

\subsection{Software Product Quality}
\begin{figure}
  \centering
  \includegraphics[width=\linewidth]{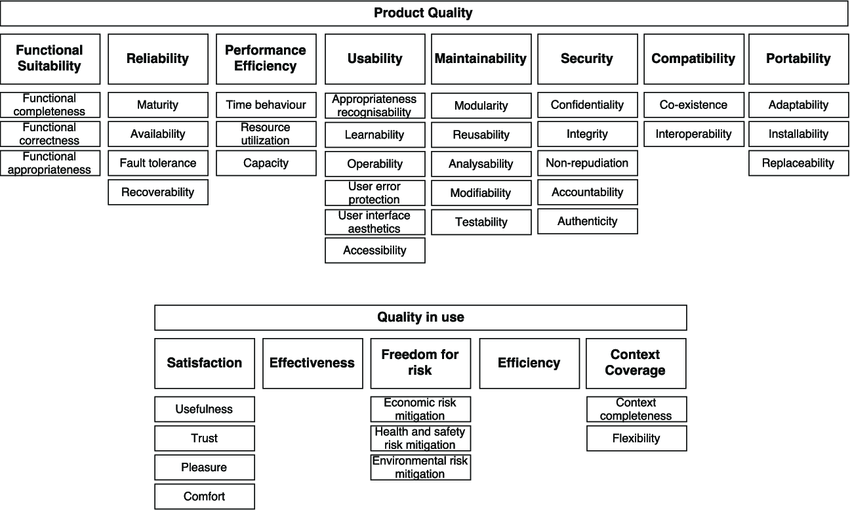}
  \caption{ISO25000 quality model \cite{ISO}}
  \label{fig:ISO}
\end{figure}

In 1991, the International Standards Organisation (ISO), issues the first version of what was then called "Software Engineering - Product Quality". It was a standard for the evaluation of software quality. ISO/IEC 9126, as this first standard was called, tried to standardize this quality evaluation by defining a quality model with standard quality characteristics and a set of metrics to evaluate these quality characteristics in software products. Over the years this standard has evolved into the ISO25000 quality model, see Figure \ref{fig:ISO}. 

Although not all software engineers might be actively using such standards or quality models, they have at least achieved that the software engineering community has a standard vocabulary while discussing software product quality and defining quality requirements. Many tools and techniques have been developed to address and evaluate the quality characteristics of a software product. 

\subsection{Trustworthy AI Systems}
It is important to realize that AI systems are also software systems, and thus the quality characteristics of ISO25000 also apply to AI systems. However, the metrics used for evaluating them might be a bit different, due to the specific nature of AI systems. One of the important differences is that AI systems not only contain code and databases, but also involve models and datasets used for training those models. Thus, for an AI system, the software product is code+data+model \cite{Sato}. All ISO25000 quality characteristics can also be applied to the dataset and the model, but the question is if there are any additional quality characteristics for AI systems. 

There are many publications about one single quality characteristics of an AI system such as transparency, bias or model performance. But we did not find many publications that contain an overview of all important quality characteristics for AI systems\footnote{Note that the ISO/IEC is working on an extension of ISO25000 for AI systems, see \url{www.iso.org/standard/80655.html}}. The EU has developed the ``Ethics Guidelines for Trustworthy AI'' \cite{EU} that contain seven key requirements:
\begin{itemize}
    \item human agency and oversight
    \item technical robustness and safety
    \item privacy and data governance
    \item transparency
    \item diversity
    \item non-discrimination and fairness
    \item environmental and societal well-being
    \item accountability
\end{itemize}
Zhang et al. \cite{zhang} look at AI systems from a testing perspective and identified nine properties to test for: correctness, overfitting degree, fairness, interpretability, robustness, security, data privacy, and efficiency.
As can be seen, there is some overlap between the EU requirements list and the testing properties from Zhang et al., but also some differences. 

\begin{figure}
  \centering
  \includegraphics[width=\linewidth]{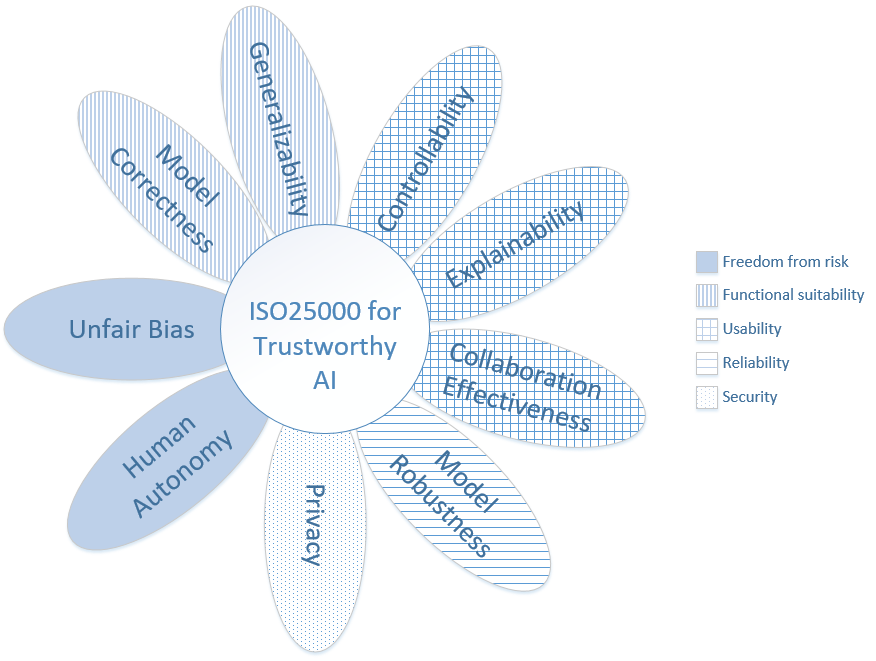}
  \caption{Extensions to ISO25000 quality model for AI systems \cite{HeckQuality}}
  \label{fig:ISOFlower}
\end{figure}

Kuwajima and Ishikawa \cite{kuwajima} analyzed the EU guidelines \cite{EU} in the light of ISO25000 to come up with six additional quality characteristics for AI systems: controllability, explainability, collaboration effectiveness, privacy, human autonomy risk mitigation and unfair bias risk mitigation. Heck \cite{HeckQuality} has combined these additions with the testing properties introduced by Zhang et al. \cite{zhang} into a set of nine additional quality characteristics for AI systems, see Figure \ref{fig:ISOFlower}. 

Since the quality model of Heck \cite{HeckQuality} is the most comprehensive one we found until now, this will be the one we will use in this paper to define quality requirements for the AI wildflower monitoring platform. 

Ahmad et al.\cite{ahmad} provide a recent overview of the state-of-practice in requirements engineering for AI systems. They found ``fewer empirical evaluations focused on ethics, explainability, and trust, and an apparent lack of research on managing data requirements and modeling requirements''. Our requirement analysis does consider ethics, explainability, and trust, for both data and model.  

\subsection{AI for Biodiversity Monitoring}
The current state-of-practice in biodiversity monitoring can be summarized as "humans - either professionals or volunteers - equipped with notebooks surveying or counting some form of wildlife in the field". Manual counting is labour-intensive, and error prone. With modern techniques such as AI and remote sensing biodiversity monitoring can be upgraded significantly \cite{klein}. Major technological breakthroughs in the last decade, especially in the field of deep learning \cite{krizhevsky2017imagenet} \cite{he2016deep} \cite{simonyan2014very} \cite{szegedy2016rethinking}, accelerated the use of AI for automatic analysis of large collections of sound and/or image data in many application domains. Also biodiversity monitoring benefits from this. Advantages of AI for biodiversity monitoring are: (1) it enables cost-effectively monitoring at a much larger (landscape) scale, and (2) the inventory results are person-independent, i.e. it is obtained with a standardized method. \flowerpower demonstrates the potential of AI for one specific form of biodiversity monitoring, viz. automatic wildflower monitoring. \footnote{Especially counting wildflowers manually is not very popular among volunteers, e.g. compared to counting birds manually.}

Previous work on AI and image-based  wildflower recognition focused largely on reliable species identification \cite{nilsback2008automated} \cite{seeland2017plant} \cite{waldchen2018automated}. Typical computer vision challenges - such as heavily cluttered backgrounds, occlusion of objects (small flowering plants are often partly covered by larger ones in patches with high vegetation), small inter-class variances (different flowering plants might be quite similar to one another) as well as large intra-class variances (individuals of the same species may vary considerably) are gradually solved over time. 

The use case of automatic wildflower monitoring with deep learning, i.e. \textit{identify} and \textit{count}, is only addressed recently. It requires an AI model that is able to both \textit{classify} and \textit{localize} objects in images. Object detection is a mature technology \cite{ren2015faster} \cite{zou2019object} already widely used in application domains as medical imaging, autonomous driving, video surveillance, robot vision, etc. Object detection models are trained with images containing tagged bounding boxes around objects of interest. Hicks et al. \cite{hicks2021deep} presented a first object detection model that was able to count 25 different wildflower species in images. In \flowerpower, the concept of Hicks et al. \cite{hicks2021deep} is expanded to a more advanced data-centric object detection model that can be used to count 100+ different wildflower species and a comprehensive software platform that supports various end-user scenarios. 

\section{Case Study Description: A Deep Learning Platform for Monitoring Wildflowers}\label{sec:case}
\flowerpower aims to build a deep learning platform for identifying and counting wildflowers. This section briefly explains the platform from a user perspective and introduces the high-level architecture of the platform. 
\subsection{User Scenarios} \label{sec:scenario}
The three most important scenarios this platform should support are detailed in this section. The scenarios are described in the form of short stories from the perspective of three different types of users: urban planners, biology researchers and AI engineers.    

\subsubsection{Scenario \#1: Citizen Science Project}
I am Anna, an urban planner responsible for urban greening policy. Our municipality has a limited budget for greening the city. Usually, it will be spent on planned park maintenance, seed mixtures for sowing wildflowers in roadsides, or excavations of new ponds at the city outskirts to provide more breeding grounds for relict amphibian populations. This year our budget was raised and for the extra money I have a different project in mind. I want to start a campaign to encourage residents of our city to participate in a citizen science project. More specifically, I want to motivate them to take smartphone photos of 1m2 patches with wildflowers in roadsides. An app will be supplied to upload these photos and take care of identifying and counting the various wildflowers automatically with AI. The collective results will be projected on an interactive city map. Also, other graphs and real-time statistics will be provided. Besides engaging our residents, persuasive visualizations for policy makers are necessary to get the important topic of urban biodiversity higher on the agenda.

\subsubsection{Scenario \#2: Systematic Biodiversity Research} 
I am Martin, a biology researcher, and a big believer of AI. This technology enlarges the impact of my research by its ability to collect data more efficiently and at larger scale, but also by its power to largely automate the analysis part. My PhD is about nature-inclusive agriculture, and the main research question is which conservation methods are most beneficial for biodiversity and at the same time economically feasible. I use wildflower richness as an indicator for biodiversity and want to systematically sample and compare wildflower richness in meadows with and without EU-funded conservation schemes for several years. The identification and counting of wildflowers will be done from aerial photos captured with a drone using an object detection model. However, since the data is collected in another region (and by another device/method), it is expected that an extension of the existing AI-model is necessary. Hence, I need tools to efficiently correct both misclassifications and mislocalizations of the current AI-model, add new wildflower species that are not yet recognized, and trigger a re-train process with an extended dataset. On top of that, I need a dashboard to show trends in flower richness over time in the selected meadows, as estimated by the latest model.

\subsubsection{Scenario \#3: Updating AI Models} 
I am Kim, an AI engineer intrigued by challenging computer vision problems and passionate about linking AI to sustainable development goals, especially themes related to climate change and biodiversity. I am very motivated to develop better solutions for identifying and counting flowering plants ‘in the wild’, with this base model as a starting point. I need functionality to upload new algorithms (preferably Jupyter notebooks), resources to easily connect them to existing high-quality annotated wildflower datasets, train these models, and compare them in a Kaggle-like leaderboard. I also want to have a nice configurable dashboard showing the detailed metrics of the newly developed algorithms, such as average precision and confusion matrices. And last but not least I also want to be able to merge and split datasets or enrich them with other data sources.

\begin{figure}
  \centering
  \includegraphics[width=\linewidth]{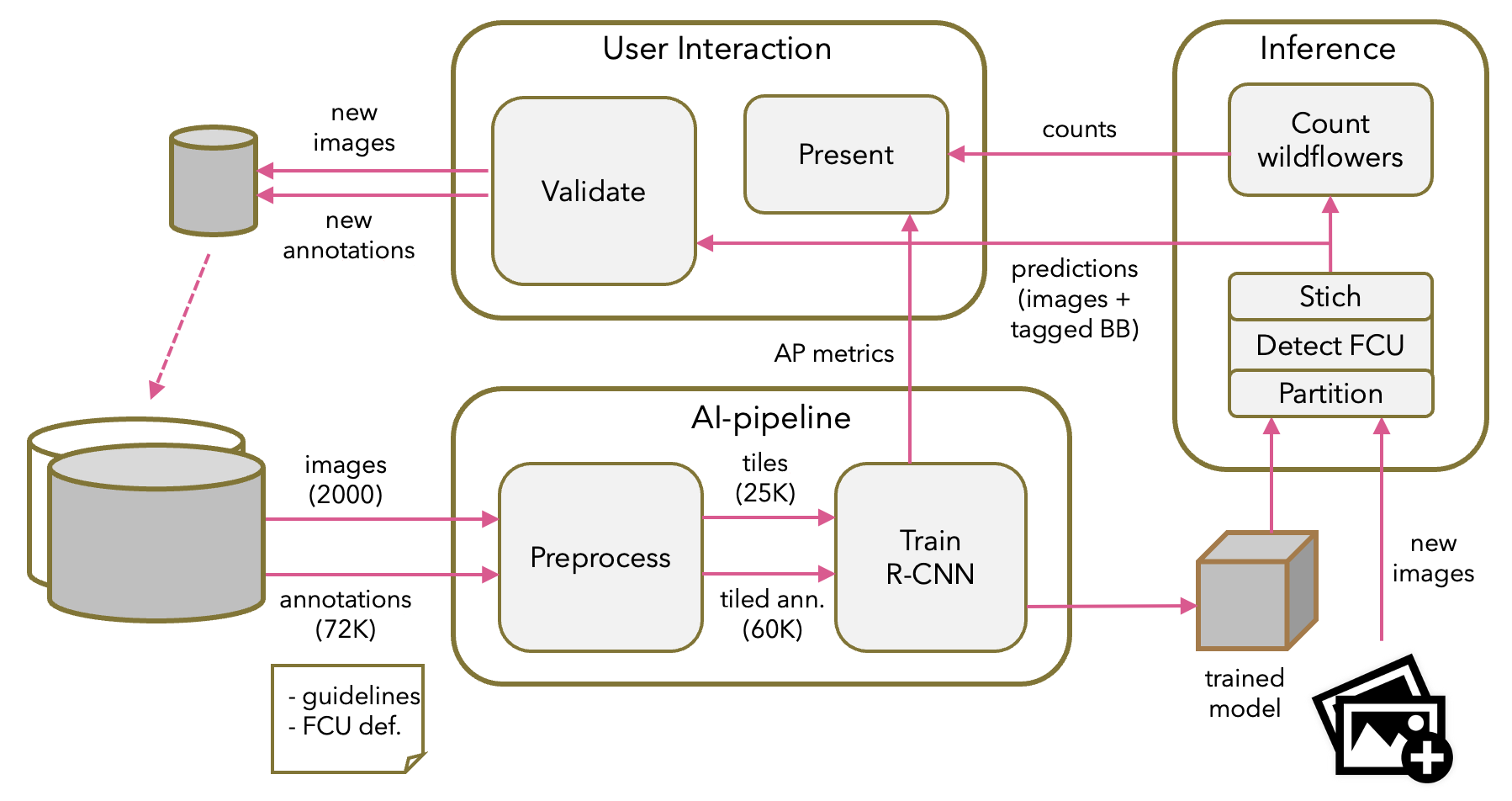}
  \caption{High-level architecture of the AI-enabled wildflower monitoring platform}
  \label{fig:architecture}
\end{figure} 

\subsection{Platform Architecture}
As shown in Figure \ref{fig:architecture} the architecture of the wildflower monitoring platform consists of three modules, that each have two components. 

The module that encapsulates the AI-pipeline is responsible for data cleaning and for training an object detection model. It is bootstrapped with 2000 images (pixel information and metadata such as GPS location of the photo as well as date and time) that are expert annotated with 72K tagged bounding boxes. 

The preprocessing component removes wildflower species (and their associated images) with less than a configurable number of annotations (set at N=50). Furthermore, it performs a stratified train-test split and partitions the high-resolution images. Usually, images are resized to squeeze them into neural networks. Partitioning makes it possible to offer native resolution tiles to the network, thereby preserving details – of wildflowers in our case – that might be crucial for identification. 

The resulting 25K tiles and their annotations are used for training and testing a Faster R-CNN object detection model (Ren et al., 2015). The trained model as well as average precision (AP) metrics on the test set (0.71 @ IoU=0.5) are exported. 

The Inference module uses the exported trained model for new wildflower images and produces predictions. The output consists of images with proposed bounding boxes around identified floral count units, and tabular wildflower counts, as estimated from the new images. 

The User Interaction module takes care of the smooth and efficient interplay with different stakeholders. It holds modules for presenting predictions and metrics (such as dashboards, maps, and leaderboards) as well as tools for validating proposed predictions and can output new images with validated annotations. 

\subsection{A Trustworthy AI-Enabled Wildflower Monitoring Platform}
Currently, several model prototypes, an annotation tool (with annotation guidelines), a data platform and a simple user interface for presenting predictions have been developed. In the coming year, the project will be evolve towards a production-ready wildflower monitoring platform along the three scenarios introduced in Section \ref{sec:scenario}. This entails improving the AI model but also a lot of software engineering as can be seen in Figure \ref{fig:architecture}. This paper focuses on the AI-related trustworthiness of the monitoring platform, by defining AI-related quality requirements for the platform. The next section details these requirements based on the quality model of Heck \cite{HeckQuality}, see Figure \ref{fig:ISOFlower}.

\section{AI-Related Quality Requirements for the Wildflower Monitoring Platform}\label{sec:req}
To build a trustworthy AI system for wildflower monitoring it is important to first analyze the quality characteristics that play a role in this system to come to AI-related quality requirements (labeled with [Qxx] below). This section describes this analysis for the nine quality characteristics in Figure \ref{fig:ISOFlower}. The requirements have been specified using the EARS syntax\cite{EARS} where applicable.

\subsection{Generalizability}
\noindent\textit{[Definition] the rules that the model learns from the training data shall also hold for unseen data}

The current wildflower object detection model of \flowerpower has been trained with about 2000 photos of about 100+ species. The images are captured with one camera type
in five different landscape types. It does of course not generalize to wildflower species that it has not been trained with. Generalizability is an important aspect when using the system in the wild. It should detect all wildflowers in the photo, indicate the species for those wildflowers that are known and mark the wildflowers that are not known as likely ``candidate'' in such a way that later-on annotations for these new species can be easily added to the system (e.g., before re-training the model). To determine what is a ``known species'' a threshold should be decided upon, e.g. prediction with above 75\% certainty. 

The model should also generalizes to photos taken with different camera types, in different resolutions and with different backgrounds. For the citizen science scenario \#1 it is to be expected that many different types of smart phones or cameras will be used. The current version of the model requires that the minimum resolution for the input images is set to 15Mpx. This ensures that details in flowers - such as stamens, carpets, petals, and sepals - are preserved in photos. For many species these details are crucial for species identification from a taxonomic perspective. It can be expected that they should be available for the object detection model as well for a correct identification. The background depends on the landscape the wildflowers have been photographed in and the weather conditions (e.g. snow or ripe). The project limits itself to open landscape types (e.g. no forest) under all weather conditions.  

\begin{description}
 \item[{[Q1]}] {When a user uploads a photo containing wildflowers, the system shall indicate for each wildflower if it is a known species or an unknown species.}
 \item[{[Q2]}] {When a user uploads a photo containing wildflowers of known species, the system shall indicate this species.}
 \item[{[Q3]}] {The system shall function with photos taken with different camera types with a resolution higher than 15Mpx.}
 \item[{[Q4]}] {The system shall function with photos taken in different open landscape types under different weather conditions.}
\end{description}

\subsection{Model Correctness}
\noindent\textit{[Definition] the rules that the model learns from the training data shall be correct}

The threshold for detecting ``known species'' from the previous paragraph also leads to a threshold for model correctness. For known species the model should classify all wildflowers of that species correctly with a certainty of at least 0.75. For unknown species, the model should classify all wildflowers as unknown (or with a certainty below 0.75 for any of the known species). 

In the wild you also see flowers that cover other flowers, damaged flowers (e.g. leaves are missing) or flowers that are still closed (buds). Hence a limit for the size of the object for it to be detected as a wildflower needs to be set. In \flowerpower the following rule is adopted: annotate and detect a blooming flower with a size of at least 1cm, at least visible for 50 \%. Furthermore it is important that there are as little false positives as possible: if the system detects a wildflower, it should also be a wildflower (and not litter or a stone for example). 

\begin{description}
 \item[{[Q5]}] {The system shall only detect flowers in there blooming phase that are at least 50\% visible and with a flower head size of at least 1cm.}
 \item[{[Q6]}] {The system shall detect all wildflowers of known species with more than 0.75 certainty.}
 \item[{[Q7]}] {Wildflowers of unknown species shall have a certainty of less than 0.75 for all known species.}
\end{description}

\subsection{Model Robustness}
\noindent\textit{[Definition] the model shall be resilient against invalid or perturbed inputs}

This project currently takes place in a controlled environment so it is unlikely that somebody will try to attack the model with e.g. adversarial photos. Therefor this quality characteristic is deemed less important for the project and does not lead to additional quality requirements. However, once the project opens up to a wider public (e.g. through the citizen science app), this quality characteristic definitely needs to be reconsidered. We know of many examples where citizens try to game the system be uploading pictures of totally unrelated objects to see how the system reacts.   

\subsection{Controllability}
\noindent\textit{[Definition] the user shall be able to control the outcome or decision of the AI system}

There are two important ways for a user to control the output of the system: 1) correct mispredicted wildflowers (bounding box or label), and 2) feed new wildflower species into the system. Especially for the systematic biodiversity research scenario \#2 these two user controls are imperative. In the ideal situation a correction or addition by the user leads to immediate retraining of the model, such that changes are directly available to the end user. Initial experiments determined that in order to retrain the model for new species, at least five images are needed with at least 50 to 80 bounding boxes with blooming wildflowers of that species. For scenario \#3 the system should support an upload function for new object detection models adhering to the input and output constraints that hold for these type of models~\cite{zou2019object}.

\begin{description}
 \item[{[Q8]}] {The user shall be able to add bounding boxes or correct the location and size of bounding boxes as predicted by model.}
 \item[{[Q9]}] {The user shall be able to add or correct the detected wildflower species (i.e. the label of the bounding box).}
 \item[{[Q10]}] {The user shall be able to request a retraining of the model with an updated or new dataset.}
  \item[{[Q11]}] {The AI engineer shall be able to upload a new object detection model and compare its metrics with the performance metrics of previous models.}
\end{description}

\subsection{Explainability}
\noindent\textit{[Definition] the system shall explain to the user on which grounds it determines its outputs}

In the annotation process some species have been aggregated. In the wild they are for example different in odour or in the detailed structure they have on there leaves but the same in their flower heads. Since the current model is only trained on flower heads, there are no distinguishing features for these specific species. The user should be aware of the limitations of the system in this sense. 

As indicated in the previous paragraph the user should be able to correct the predictions of the system. From a data science perspective it is also important to understand why the system has mispredicted. Is the object not a wildflower? Is the bounding box to small or large? Do two species look alike? The AI engineer mentioned in scenario \#3 could greatly be helped by the user if he/she provides feedback of what went wrong. For the user to provide useful feedback about what went wrong, the system should provide as much information as possible about the prediction (not only the end result, but also e.g. a top-5 of predicted species with their certainty). 

An AI engineer developing next generation object detect models needs more advanced techniques for explainable AI \cite{xai}, such as so-called saliency maps, i.e. heatmap-like image overlays indicating how much each pixel contributes to the prediction \cite{petsiuk2021black}.

\begin{description}
 \item[{[Q12]}]{When wildflower species have been aggregated, the system shall alert the user for this and explain why this has been done.}
 \item[{[Q13]}] {When bounding boxes or wildflower species have been mispredicted, the system shall offer the user the possibility to provide feedback about the nature and suspected reason of the mispredictions.}
 \item[{[Q14]}] {While presenting predictions to the user, the system shall provide a top-5 of predicted wildflower species and their certainty per bounding box.}
 \item[{[{Q15}]}] {While presenting predictions to the user, the system shall also provide saliency maps.}
\end{description}

\subsection{Collaboration Effectiveness}
\noindent\textit{[Definition] the system shall effectively collaborate with the user in accomplishing the required tasks}

For the system to be useful to the biology researchers it should be able to function in two different modes: 1) select a batch of photos from one location and count the total number of wildflowers per species, 2) select one single photo and detect (bounding boxes) individual wildflowers and their species (labels). As said when discussing Controllability, it would be very effective if the system automatically retrains after corrections, so the updated model with the new species can immediately be used by the user for new predictions, but at least Q10 should hold. 

\begin{description}
 \item[{[Q16]}]{The system shall have a batch mode to aggregate wildflower species counts over multiple photos.}
\end{description}

\subsection{Privacy}
\noindent\textit{[Definition] the dataset shall be free from personal data or data that can indirectly trace back to individual persons}

Since the main object in the system is wildflowers, it may seem that privacy is not an issue. But it is still humans that are taking the photos, uploading them and annotating them. So the system must ensure that the privacy of these persons is protected as much as possible, especially in the citizen science scenario \#1. For example, metadata that is stored with every photo like camera type or GPS location, could lead to identifying the exact person that took the photo and thus also some private information about this person like the fact that he or she has a very expensive camera and walks in area XYZ every Thursday afternoon. 

On the other hand, the biology researchers stated that not only the privacy of people is important, but also protection of wildlife (in this case ``the privacy'' of wildflowers). If the system for example holds the photo of a very rare orchid, they would like to obfuscate the exact location of that orchid, to prevent the general public from going there and perhaps destroying vulnerable species. 

\begin{description}
 \item[{[Q17]}]{The system shall remove persons or objects that could identify specific persons from the photos.}
 \item[{[Q18]}]{The system shall remove or obfuscate person-related metadata from the photos.}
 \item[{[Q19]}]{The system shall protect the exact location of rare and vulnerable wildflowers.}
\end{description}

\subsection{Unfair Bias}
\noindent\textit{[Definition] the rules that the model learns are not unfair to or biased against a certain group of the population}

One would say that the system is not in any way going to affect the lives of people by treating them unfair, because it is ``just about wildflowers''. But the same as for privacy (see previous paragraph) the system could also treat wildflowers unfair. If certain rare species are neglected in the training set, they will never be detected in new photos that are uploaded. In the same way one could reason that if certain types or locations of landscapes are neglected in the training dataset, their biodiversity will not be monitored, in the end affecting the lives of people living near those neglected areas. 

As said before, the current model needs at least 50 to 80 bounding boxes of a wildflower species to be able to train the model for reliably detecting that species. Thus, the system should ensure that there is a mechanism for alerting or removing underrepresented species, with less annotations than a certain threshold.

\begin{description}
 \item[{[Q20]}]{The system shall be trained with a well-balanced dataset with respect to species occurrence, locations and landscape types.}
  \item[{[Q21]}]{The system shall have a configurable parameter removing annotations (as well as the associated images on which these annotations occur) of underrepresented species.}
\end{description}

\subsection{Human Autonomy}
\noindent\textit{[Definition] the system shall not threaten users in their autonomy}

Acceptance of AI solutions - although often problematic in many disciplines - goes fast and smooth in biology research. Biology researchers are very much in need of automated annotation and monitoring solutions, because currently it is very hard to recruit volunteers to do this manually. AI also extends the impact of biodiversity research by enabling monitoring on a landscape scale. Having an automated solution for counting in the wild actually increases the feeling of human autonomy since the biology researchers can then more focus on innovative projects in stead of doing the arduous job of counting wildflowers. The citizen science scenario \#1, raises more awareness for nature by educating the general public about the wildflower richness that can be found in their environment, thus increasing the human autonomy of the general public. Therefor this quality characteristic does not lead to additional quality requirements for this project.  

\section{Discussion}\label{sec:discuss}
This section discusses the above analysis in light of the future developments of the \flowerpowersystem. It also summarizes the lessons learned for other projects developing production-ready AI systems. 

\subsection{Future of \flowerpower}
The above analysis only considered the nine AI-related quality characteristics depicted in Figure \ref{fig:ISOFlower}. The existing ISO25000 quality characteristics also apply to the \flowerpowersystem. So, the analysis should be elaborated with all characteristics as depicted in Figure \ref{fig:ISO}, to collect a complete list of quality requirements for the \flowerpowersystem. It is also important to prioritize this list, since there is always a cost-quality trade-off. For example, implementing MLOps \cite{GoogleMLOps} is planned for the \flowerpowersystem, but it is still to be investigated if the system needs the continuous training of the higher MLOps levels or if models could be retrained manually (see [Q10] above).

The analysis of quality requirements at such an early stage also resulted in a better view on the testing of the system. For each of the 21 quality requirements we have asked ourselves how to test them. This led to the observation that system needs to be tested with different types of images, e.g. with small objects that look like wildflowers [Q5], from different types of landscapes [Q4], with different camera types and resolutions [Q3] and with persons or their personal stuff in the photo [Q17]. Collecting such images should start as early as possible. Other requirements are more easy to check, as they indicate desired functionality: correcting predictions [Q8 till Q10], upload new models [Q11], alert for aggregations [Q12], provide feedback [Q13], top-5 predictions [Q14], saliency maps [Q15], batch mode [Q16], remove meta data [Q18], remove exact location [Q19], remove underrepresented species [Q21]. Also, the project should keep investing a collecting new data and extending its already high-quality dataset [Q1 till 7, Q20]. For this the project should also look into other data augmentation methods than just taking photos in the wild.

The \flowerpowersystem is planned to evolve from a Jupyter Notebook and a separate annotation tool, into a full-fledged AI system supporting biology researchers and other stakeholders in monitoring wildflowers. This project treats data, models and code as first-class citizens, as it deems the quality of all three equally important for the quality of the overall system. A next step is to use ISO25000 and the extension proposed by Heck \cite{HeckQuality} to define quality requirements for data, model and software in a full-fledged requirements engineering process with all stakeholders involved as a solid base for the further AI engineering of the system.  

\subsection{Requirements Engineering for AI Systems}
\begin{figure}
  \centering
  \includegraphics[width=\linewidth]{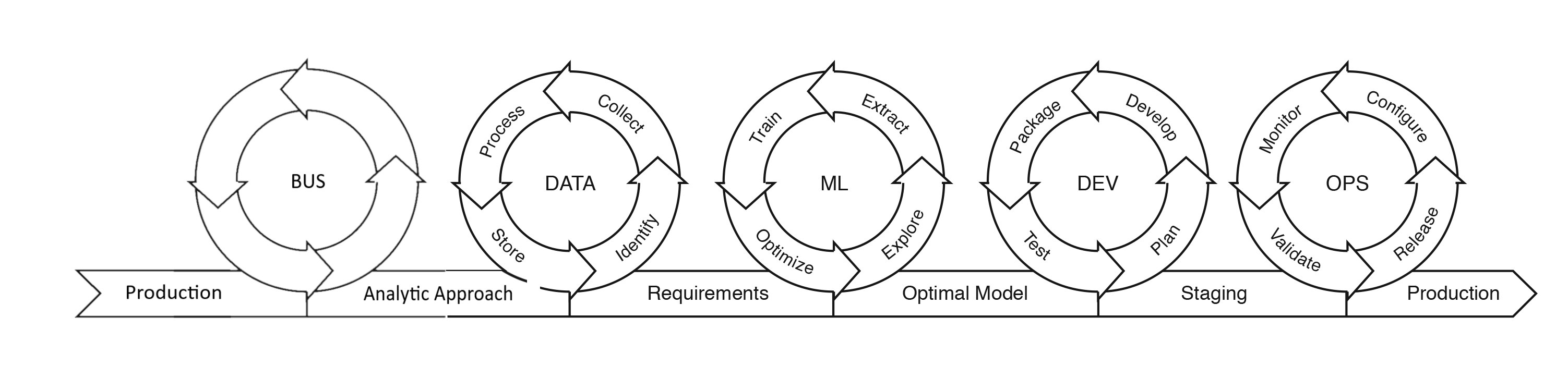}
  \caption{Extended AI engineering process \cite{Heck2022}}
  \label{fig:BU}
\end{figure}

As stated by Ahmad et al.\cite{ahmad}, the data and model requirements are often overlooked in AI engineering. This matches our observation from previous work \cite{Heck2022} that the traditional process depicted for AI engineering should also include a ``Business Understanding (BUS)'' phase as present in the CRISP-DM model\cite{CRISPDM}, see Figure \ref{fig:BU}. This business understanding phase is a pivotal step in building production-ready AI systems. It is the step where the project should involve all stakeholders (including data engineers, data scientists and devops engineers) to define the requirements, including those for the data and the model. Also in the FlowerPower project we observed that it is essential to define requirements for all aspects of the platform from the start of the project. As can be seen in the previous paragraph, the AI-related requirements gave us pointers what data to collect or how to build our models. 

\subsection{A Quality Model for Trustworthy AI Systems}
Analyzing the \flowerpowersystem along the nine quality properties as proposed by Heck \cite{HeckQuality} gave many insights into the requirements of a trustworthy AI system for monitoring wildflowers. It provided a way to translate the high-level concept of trustworthiness into more concrete and measurable quality requirements. Of course, this was already the case for ISO25000, but our analysis shows that these nine additional characteristics are indeed a valuable addition when analyzing the quality of AI systems.    
The list of quality requirements of the \flowerpowersystem includes requirements about all three components of AI systems:
\begin{itemize}
    \item data: [Q3], [Q4], [Q17] till [Q21]
    \item model: [Q1], [Q2], [Q5] till [Q7], [Q10], [Q11]
    \item software: [Q8], [Q9], [Q12] till [Q16]
\end{itemize}  
That means that when analyzing quality properties of an AI system, indeed all three aspects should be considered. Also when applying the ISO25000 list of characteristics. For example, the time behaviour characteristic should also consider the timing-related performance of the model and the resource utilization characteristic should also consider the storage of training data.  
In our opinion it helps to have a standard dictionary for quality requirements for AI systems, just like ISO25000 functions for software products. For each of the nine characteristics there is a whole body of knowledge (and tools) being built up, but they do not always use the same terminology. A next step is to uncover tools, techniques, methods and frameworks that aid in achieving or measuring the desired quality characteristics. The ultimate goal is to define a quality engineering toolbox for AI systems that can be used in any AI engineering project. Collecting all this related work per characteristic and translating this to coherent guidelines and tools for practitioners building AI systems could hugely help in maturing the discipline of AI engineering.     

\section{Conclusion}
This paper applies the quality model for trustworthy AI systems as introduced by Heck \cite{HeckQuality} to define AI-related quality requirements for a real-life AI system, called \flowerpower. Although \flowerpower is a custom-developed platform for monitoring wildflowers, it has many properties that hold for other AI systems as well.

This paper demonstrates how a quality model can be used a standard dictionary for defining quality requirements for AI systems. Defining and implementing such quality requirements is a pivotal step in going from a research prototype or trained model to a trustworthy AI system in a production environment. An integrated AI engineering approach for trustworthy AI systems should use the quality model for defining quality requirements for all aspects of the system: data, models and code. 

Ongoing work is to extend the quality model with metrics, tools and best practices to aid AI engineers in implementing trustworthy AI systems with the relevant quality characteristics. 

\section*{Acknowledgment}
We would like to thank \fontys and Naturalis Biodiversity Center for supporting our research. 

\balance
\bibliographystyle{ieeetr}
\bibliography{library, petra, Gerard}

\begin{thebibliography}{10}

\bibitem{ipbes}
IPBES, ``{Global assessment report on biodiversity and ecosystem services of
  the Intergovernmental Science-Policy Platform on Biodiversity and Ecosystem
  Services},'' May 2019.

\bibitem{klein}
D.~J. Klein, M.~W. McKown, and B.~R. Tershy, ``Deep learning for large scale
  biodiversity monitoring,'' in {\em Bloomberg Data for Good Exchange
  Conference}, 2015.

\bibitem{Bosch2021}
J.~Bosch, H.~H. Olsson, and I.~Crnkovic, ``Engineering ai systems: A research
  agenda,'' {\em Artificial Intelligence Paradigms for Smart Cyber-Physical
  Systems}, pp.~1--19, 2021.

\bibitem{EU}
H.-L. E.~G. on~Artificial Intelligence (AI~HLEG), ``Ethics guidelines for
  trustworthy ai.'' online, 2018.

\bibitem{ISO}
L.~M.~G. Rodr{\'\i}guez, F.~Oquendo, E.~Y. Nakagawa, {\em et~al.}, ``Qm4aal:
  quality model for ambient assisted living systems.,'' 2017.

\bibitem{Sato}
D.~Sato, A.~Wider, and C.~Windheuser, ``Continuous delivery for machine
  learning.'' online, 2019.

\bibitem{zhang}
J.~M. Zhang, M.~Harman, L.~Ma, and Y.~Liu, ``Machine learning testing: Survey,
  landscapes and horizons,'' {\em IEEE Transactions on Software Engineering},
  2020.

\bibitem{kuwajima}
H.~Kuwajima and F.~Ishikawa, ``Adapting square for quality assessment of
  artificial intelligence systems,'' in {\em 2019 IEEE International Symposium
  on Software Reliability Engineering Workshops (ISSREW)}, pp.~13--18, IEEE,
  2019.

\bibitem{HeckQuality}
P.~Heck, ``A quality model for trustworthy ai systems.'' online, 2021.

\bibitem{ahmad}
K.~Ahmad, M.~Abdelrazek, C.~Arora, M.~Bano, and J.~Grundy, ``Requirements
  engineering for artificial intelligence systems: A systematic mapping
  study,'' {\em Information and Software Technology}, p.~107176, 2023.

\bibitem{krizhevsky2017imagenet}
A.~Krizhevsky, I.~Sutskever, and G.~E. Hinton, ``Imagenet classification with
  deep convolutional neural networks,'' {\em Communications of the ACM},
  vol.~60, no.~6, pp.~84--90, 2017.

\bibitem{he2016deep}
K.~He, X.~Zhang, S.~Ren, and J.~Sun, ``Deep residual learning for image
  recognition,'' in {\em Proceedings of the IEEE conference on computer vision
  and pattern recognition}, pp.~770--778, 2016.

\bibitem{simonyan2014very}
K.~Simonyan and A.~Zisserman, ``Very deep convolutional networks for
  large-scale image recognition,'' {\em arXiv preprint arXiv:1409.1556}, 2014.

\bibitem{szegedy2016rethinking}
C.~Szegedy, V.~Vanhoucke, S.~Ioffe, J.~Shlens, and Z.~Wojna, ``Rethinking the
  inception architecture for computer vision,'' in {\em Proceedings of the IEEE
  conference on computer vision and pattern recognition}, pp.~2818--2826, 2016.

\bibitem{nilsback2008automated}
M.-E. Nilsback and A.~Zisserman, ``Automated flower classification over a large
  number of classes,'' in {\em 2008 Sixth Indian Conference on Computer Vision,
  Graphics \& Image Processing}, pp.~722--729, IEEE, 2008.

\bibitem{seeland2017plant}
M.~Seeland, M.~Rzanny, N.~Alaqraa, J.~W{\"a}ldchen, and P.~M{\"a}der, ``Plant
  species classification using flower images—a comparative study of local
  feature representations,'' {\em PloS one}, vol.~12, no.~2, p.~e0170629, 2017.

\bibitem{waldchen2018automated}
J.~W{\"a}ldchen, M.~Rzanny, M.~Seeland, and P.~M{\"a}der, ``Automated plant
  species identification—trends and future directions,'' {\em PLoS
  computational biology}, vol.~14, no.~4, p.~e1005993, 2018.

\bibitem{ren2015faster}
S.~Ren, K.~He, R.~Girshick, and J.~Sun, ``Faster r-cnn: Towards real-time
  object detection with region proposal networks,'' {\em Advances in neural
  information processing systems}, vol.~28, 2015.

\bibitem{zou2019object}
Z.~Zou, Z.~Shi, Y.~Guo, and J.~Ye, ``Object detection in 20 years: A survey,''
  {\em arXiv preprint arXiv:1905.05055}, 2019.

\bibitem{hicks2021deep}
D.~Hicks, M.~Baude, C.~Kratz, P.~Ouvrard, and G.~Stone, ``Deep learning object
  detection to estimate the nectar sugar mass of flowering vegetation,'' {\em
  Ecological Solutions and Evidence}, vol.~2, no.~3, p.~e12099, 2021.

\bibitem{EARS}
A.~Mavin, P.~Wilkinson, A.~Harwood, and M.~Novak, ``Easy approach to
  requirements syntax (ears),'' in {\em 2009 17th IEEE International
  Requirements Engineering Conference}, pp.~317--322, IEEE, 2009.

\bibitem{xai}
W.~Samek, G.~Montavon, A.~Vedaldi, L.~K. Hansen, and K.-R. M{\"u}ller, {\em
  Explainable AI: interpreting, explaining and visualizing deep learning},
  vol.~11700.
\newblock Springer Nature, 2019.

\bibitem{petsiuk2021black}
V.~Petsiuk, R.~Jain, V.~Manjunatha, V.~I. Morariu, A.~Mehra, V.~Ordonez, and
  K.~Saenko, ``Black-box explanation of object detectors via saliency maps,''
  in {\em Proceedings of the IEEE/CVF Conference on Computer Vision and Pattern
  Recognition}, pp.~11443--11452, 2021.

\bibitem{GoogleMLOps}
Google, ``Mlops: Continuous delivery and automation pipelines in machine
  learning.'' online.

\bibitem{Heck2022}
M.~Meesters, P.~Heck, and A.~Serebrenik, ``What is an ai engineer? an empirical
  analysis of job ads in the netherlands,'' in {\em Proceedings of the 1st
  International Conference on AI Engineering: Software Engineering for AI},
  pp.~136--144, 2022.

\bibitem{CRISPDM}
P.~Chapman, J.~Clinton, R.~Kerber, T.~Khabaza, T.~Reinartz, C.~Shearer, and
  R.~Wirth, ``Crisp-dm 1.0 step-by-step data mining guides,'' 2000.

\end{thebibliography}

\end{document}